\documentclass[11pt]{article}
\usepackage{float}
\usepackage{acl}
\usepackage{times}
\usepackage{latexsym}
\usepackage{amsfonts,amssymb}
\usepackage{amsmath}
\usepackage{graphicx}
\usepackage{float}
\usepackage{array}
\usepackage{multirow}
\usepackage{url}
\usepackage{stfloats}

\aclfinalcopy
\title{TransG : A Generative Model for Knowledge Graph Embedding}
\title{TransG : A Generative Model for Knowledge Graph Embedding}
\author{Han Xiao$^*$, Minlie Huang$^*$, Xiaoyan Zhu \\
	State Key Lab. of Intelligent Technology and Systems, \\
	National Lab. for Information Science and Technology, \\ Dept. of Computer Science and Technology, Tsinghua University, Beijing 100084, PR China
	\\ bookman@vip.163.com; \{aihuang,zxy-dcs\}@tsinghua.edu.cn \\
	$^*$Corresponding Authors: \url{http://www.ibookman.net}, \url{http://www.aihuang.org}
}

\date{}

\begin{document}

\maketitle

\begin{abstract}
	Recently, knowledge graph embedding, which projects symbolic entities and relations into continuous vector space, has become a new, hot topic in artificial intelligence. This paper proposes a novel generative model (\textbf{TransG}) to address the issue of \textbf{multiple relation semantics} that a relation may have multiple meanings revealed by the entity pairs associated with the corresponding triples. The new model can discover latent semantics for a relation and leverage a mixture of relation-specific component vectors to embed a fact triple. To the best of our knowledge, this is the first generative model for knowledge graph embedding, and at the first time, the issue of multiple relation semantics is formally discussed. Extensive experiments show that the proposed model achieves substantial improvements against the state-of-the-art baselines. All of the related poster, slides, datasets and codes have been published in \url{http://www.ibookman.net/conference.html}.
\end{abstract}

\section{Introduction}

Abstract or real-world knowledge is always a major topic in Artificial Intelligence. Knowledge bases such as Wordnet \cite{miller1995wordnet} and Freebase \cite{bollacker2008freebase} have been shown very useful to AI tasks including question answering, knowledge inference, and so on. However, traditional knowledge bases are symbolic and logic, thus numerical machine learning methods cannot be leveraged to support the computation over the knowledge bases. To this end, knowledge graph embedding has been proposed to project entities and relations into continuous vector spaces. Among various embedding models, there is a line of translation-based models such as TransE \cite{bordes2013translating}, TransH \cite{wang2014knowledge}, TransR \cite{lin2015learning}, and other related models \cite{he2015learning} \cite{lin2015modeling}.

\begin{figure}[H]
	\centering
	\includegraphics[width=1.0\linewidth]{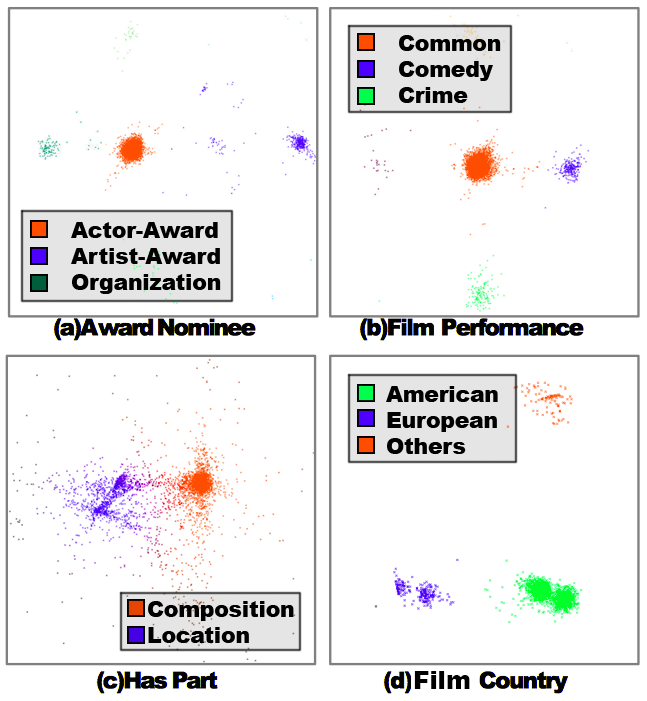}
	\caption{Visualization of TransE embedding vectors with PCA dimension reduction. Four relations (a $\thicksim$ d) are chosen from Freebase and Wordnet. A dot denotes a triple and its position is decided by the difference vector between tail and head entity ($\mathbf{t-h}$). Since TransE adopts the principle of $\mathbf{t-h \approx r}$, there is supposed to be only one cluster whose centre is the relation vector $\mathbf{r}$. However, results show that there exist multiple clusters, which justifies our multiple relation semantics assumption.}
	\label{fig:fig_1}
\end{figure}

A fact of knowledge base can usually be represented by a triple $(h,r,t)$ where $h,r,t$ indicate a head entity, a relation, and a tail entity, respectively. All translation-based models almost follow the same principle $\mathbf{h_r + r \approx t_r}$ where $\mathbf{h_r,r,t_r}$ indicate the embedding vectors of triple $(h,r,t)$, with the head and tail entity vector projected with respect to the relation space.

\begin{figure*}
	\centering
	\includegraphics[width=0.75\linewidth]{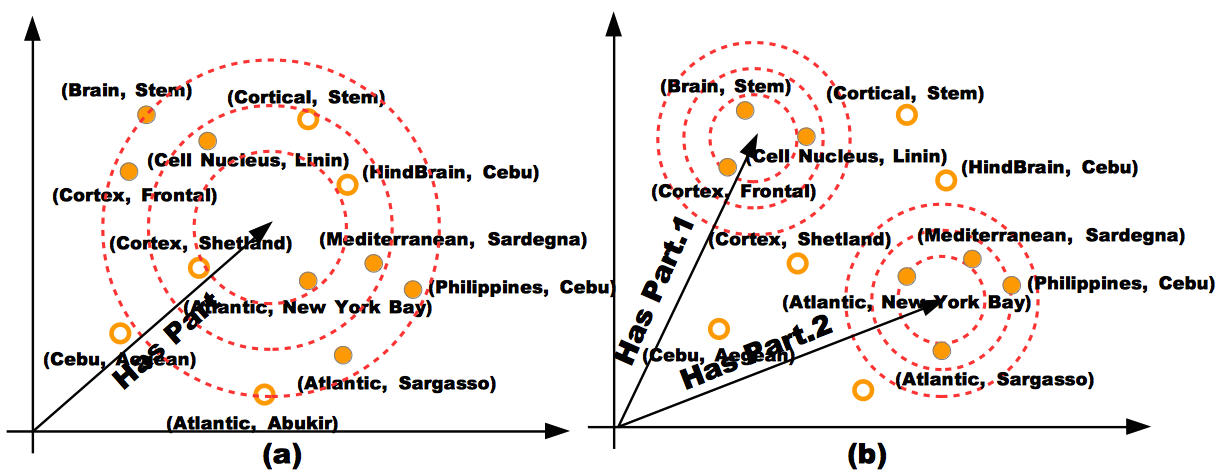}
	\caption{Visualization of multiple relation semantics. The data are selected from Wordnet. The dots are correct triples that belong to $\mathrm{HasPart}$ relation, while the circles are incorrect ones. The point coordinate is the difference vector between tail and head entity, which should be near to the centre. (a) The correct triples are hard to be distinguished from the incorrect ones. (b) By applying multiple semantic components, our proposed model could discriminate the correct triples from the wrong ones.}
	\label{fig:fig_2}
\end{figure*}

In spite of the success of these models, none of the previous models has formally discussed the issue of \textbf{\textit{multiple relation semantics}} that a relation may have multiple meanings revealed by the entity pairs associated with the corresponding triples. As can be seen from Fig.~\ref{fig:fig_1}, visualization results on embedding vectors obtained from TransE \cite{bordes2013translating} show that, there are different clusters for a specific relation, and different clusters indicate different latent semantics. For example, the relation \textrm{HasPart} has at least two latent semantics: composition-related as \textrm{(Table, HasPart, Leg)} and location-related as \textrm{(Atlantics, HasPart, NewYorkBay)}.  As one more example, in Freebase, \textrm{(Jon Snow, birth place, Winter Fall)} and \textrm{(George R. R. Martin, birth place, U.S.)} are mapped to schema \textrm{/fictional\_universe/fictional\_character/place\_of\_birth} and \textrm{/people/person/place\_of\_birth} respectively, indicating that \textit{birth place} has different meanings.
This phenomenon is quite common in knowledge bases for two reasons: artificial simplification and nature of knowledge. On one hand, knowledge base curators could not involve too many similar relations, so abstracting multiple similar relations into one specific relation is a common trick. On the other hand, both language and knowledge representations often involve ambiguous information. The ambiguity of knowledge means a semantic mixture.
For example, when we mention ``Expert'', we may refer to scientist, businessman or writer, so the concept ``Expert'' may be ambiguous in a specific situation, or generally a semantic mixture of these cases.

However, since previous translation-based models adopt $\mathbf{h_r + r \approx t_r}$, they assign only one translation vector for one relation, and these models are not able to deal with the issue of multiple relation semantics. To illustrate more clearly, as showed in Fig.\ref{fig:fig_2}, there is only one unique representation for relation $\mathrm{HasPart}$ in traditional models, thus the models made more errors when embedding the triples of the relation. Instead, in our proposed model, we leverage a Bayesian non-parametric infinite mixture model to handle multiple relation semantics by generating multiple translation components for a relation. Thus, different semantics are characterized by different components in our embedding model. For example, we can distinguish the two clusters $\mathrm{HasPart.1}$ or $\mathrm{HasPart.2}$, where the relation semantics are automatically clustered to represent the meaning of associated entity pairs.

To summarize, our contributions are as follows:
\begin{itemize}
	\item We propose a new issue in knowledge graph embedding, \textbf{\textit{multiple relation semantics}} that a relation in knowledge graph may have different meanings revealed by the associated entity pairs, which has never been studied previously.
	\item To address the above issue, we propose a novel Bayesian non-parametric infinite mixture embedding model, TransG. The model can automatically discover semantic clusters of a relation, and leverage a mixture of multiple relation components for translating an entity pair. Moreover, we present new insights from the generative perspective.
	\item Extensive experiments show that our proposed model obtains substantial improvements against the state-of-the-art baselines.
\end{itemize}


\section{Related Work}
Prior studies are classified into two branches: translation-based embedding methods and the others.

\subsection{Translation-Based Embedding Methods}
Existing translation-based embedding methods share the same translation principle $\mathbf{h + r \approx t}$ and the score function is designed as: $$f_r(h,t) = ||\mathbf{h_r + r - t_r}||_2^2$$ where $\mathbf{h_r, t_r}$ are entity embedding vectors projected in the relation-specific space. \textbf{TransE} \cite{bordes2013translating}, lays the entities in the original entity space: $\mathbf{h_r = h,  t_r = t}$. \textbf{TransH} \cite{wang2014knowledge}, projects entities into a hyperplane for addressing the issue of complex relation embedding: $\mathbf{h_r = h - w_r^\top hw_r,  t_r = t - w_r^\top t w_r}$. To address the same issue, \textbf{TransR} \cite{lin2015learning}, transforms the entity embeddings by the same relation-specific matrix: $\mathbf{h_r=M_rh, t_r=M_rt}$. TransR also proposes an ad-hoc clustering-based method, \textbf{CTransR}, where the entity pairs for a relation are clustered into different groups, and the pairs in the same group share the same relation vector. In comparison, our model is more elegant to address such an issue theoretically, and does not require a pre-process of clustering. Furthermore, our model has much better performance than CTransR, as we expect. \textbf{TransM} \cite{fan2014transition} leverages the structure of the knowledge graph via pre-calculating the distinct weight for each training triple to enhance embedding. \textbf{KG2E} \cite{he2015learning} is a probabilistic embedding method for modeling the uncertainty in knowledge graph.

\subsection{Pioneering Embedding Methods}
There list the pioneering embedding approaches:

\textbf{Structured Embedding (SE).} The SE model \cite{bordes2011learning} transforms the entity space with the head-specific and tail-specific matrices. The score function is defined as $f_r(h,t)=||\mathbf{M_{h,r}h - M_{t,r}t}||$. According to \cite{socher2013reasoning}, this model cannot capture the relationship between entities. \textbf{Semantic Matching Energy (SME).} The SME model \cite{bordes2012joint} \cite{bordes2014semantic} can handle the correlations between entities and relations by matrix product and Hadamard product.
In some recent work \cite{bordes2014semantic}, the score function is re-defined with 3-way tensors instead of matrices. \textbf{Single Layer Model (SLM).} SLM applies neural network to knowledge graph embedding. The score function is defined as $f_r(h,t)=\mathbf{u_r^\top}g(\mathbf{M_{r,1}h + M_{r,2}t}) $ where
$\mathbf{M_{r,1}, M_{r,2}}$ are relation-specific weight matrices. Collobert had applied a similar method into the language model, \cite{collobert2008unified}. \textbf{Latent Factor Model (LFM).} The LFM \cite{jenatton2012latent}, \cite{sutskever2009modelling} attempts to capture the second-order correlations between entities by a quadratic form. The score function is as $f_r(h,t) = \mathbf{h^\top W_rt}$. \textbf{Neural Tensor Network (NTN).} The NTN model \cite{socher2013reasoning} defines a very expressive score function to combine the SLM and LFM: $f_r(h,t) =\mathbf{u_r^\top}g(\mathbf{h^\top W_{\cdot \cdot r}t + M_{r,1}h + M_{r,2}t + b_r}) $, where $\mathbf{u_r}$ is a relation-specific linear layer, $g(\cdot)$ is the $tanh$ function, $\mathbf{W} \in \mathbb{R}^{d \times d \times k}$ is a 3-way tensor. \textbf{Unstructured Model (UM).} The UM \cite{bordes2012joint} may be a simplified version of TransE without considering any relation-related information. The score function is directly defined as $f_r(h,t) = ||\mathbf{h - t}||_2^2$. \textbf{RESCAL.} This is a collective matrix factorization model which is also a common method in knowledge base embedding \cite{nickel2011three}, \cite{nickel2012factorizing}. 

\textbf{Semantically Smooth Embedding (SSE).} \cite{guo2015semantically} aims at further discovering the geometric structure of the embedding space to make it semantically smooth.
\textbf{\cite{wang2014knowledge}} focuses on bridging the gap between knowledge and texts, with a joint loss function for knowledge graph and text corpus.  \textbf{\cite{wang2015knowledge}} incorporates the rules that are related with relation types such as 1-N and N-1. \textbf{PTransE.} \cite{lin2015modeling} is a path-based embedding model, simultaneously considering the information and confidence level of the path in knowledge graph.

\section{Methods}


\subsection{TransG: A Generative Model for Embedding}
As just mentioned, only one single translation vector for a relation may be insufficient to model multiple relation semantics. In this paper, we propose to use Bayesian non-parametric infinite mixture embedding model \cite{griffiths2011indian}. The generative process of the model is as follows:

\begin{enumerate}
	\item For an entity $e \in E$:
	\begin{enumerate}
		\item Draw each entity embedding mean vector from a standard normal distribution as a prior: $\mathbf{u_e \backsim \mathcal{N}(0,1)}$.
	\end{enumerate}	
	\item For a triple $(h, r, t) \in \Delta$:
	\begin{enumerate}
		\item Draw a semantic component from Chinese Restaurant Process for this relation: $\pi_{r,m} \thicksim CRP(\beta)$.
		\item Draw a head entity embedding vector from a normal distribution: $\mathbf{h} \backsim \mathcal{N}(\mathbf{u_h}, \sigma_h^2 \mathbf{E})$.
		\item Draw a tail entity embedding vector from a normal distribution: $\mathbf{t} \backsim \mathcal{N}(\mathbf{u_t}, \sigma_t^2 \mathbf{E})$.
		\item Draw a relation embedding vector for this semantics: $\mathbf{u_{r,m}} = \mathbf{t - h} \backsim \mathcal{N}(\mathbf{u_t-u_h}, (\sigma_h^2 + \sigma_t^2)\mathbf{E})$.
	\end{enumerate}
\end{enumerate}
where $\mathbf{u_h}$ and $\mathbf{u_t}$ indicate the mean embedding vector for head and tail respectively, $\sigma_h$ and $\sigma_t$ indicate the variance of corresponding entity distribution respectively, and $\mathbf{u_{r,m}}$ is the $m$-th component translation vector of relation $r$. Chinese Restaurant Process (CRP) is a Dirichlet Process and it can automatically detect semantic components. In this setting, we obtain the score function as below:
\begin{eqnarray}
& \mathbb{P}\{(h, r, t)\} & \propto \sum_{m=1}^{M_r} \pi_{r,m} \mathbb{P}(\mathbf{u_{r,m}} | h,t)  \nonumber \\ & & = \sum_{m=1}^{M_r} \pi_{r,m} e^{-\frac{||\mathbf{u_h + u_{r,m} - u_t}||_2^2}{\sigma_h^2 + \sigma_t^2}} 
\end{eqnarray}
where $\pi_{r,m}$ is the mixing factor, indicating the weight of $i$-th component and $M_r$ is the number of semantic components for the relation $r$, which is learned from the data automatically by the CRP.

Inspired by Fig.\ref{fig:fig_1}, TransG leverages a mixture of relation component vectors for a specific relation. Each component represents a specific latent meaning. By this way, TransG could distinguish multiple relation semantics. Notably, the CRP could generate multiple semantic components when it is necessary and the relation semantic component number $M_r$ is learned adaptively from the data.

\subsection{Explanation from the Geometry Perspective}
Similar to previous studies, TransG has geometric explanations. In the previous methods, when the relation $r$ of triple $(h,r,t)$ is given, the geometric representations are fixed, as $\mathbf{h + r \approx t}$. However, TransG generalizes this geometric principle to:
\begin{eqnarray}
& m_{(h,r,t)}^*  = \underset{m=1...M_r}{\arg\max} ~~ \left( \pi_{r,m} e^{-\frac{||\mathbf{u_h+u_{r,m}-u_t}||_2^2}{\sigma_h^2 + \sigma_t^2}} \right) & \nonumber \\
& \mathbf{h + u_{r,m_{(h,r,t)}^*}}\approx \mathbf{t} &
\end{eqnarray}
where $m_{(h,r,t)}^*$ is the index of primary component. Though all the components contribute to the model, the primary one contributes the most due to the exponential effect ($exp(\cdot)$). When a triple $(h,r,t)$ is given, TransG works out the index of primary component then translates the head entity to the tail one with the primary translation vector.


For most triples, there should be only one component that have significant non-zero value as $  \left( \pi_{r,m} e^{-\frac{||\mathbf{u_h+u_{r,m}-u_t}||_2^2}{\sigma_h^2 + \sigma_t^2}} \right) $ and the others would be small enough, due to the exponential decay. This property reduces the noise from the other semantic components to better characterize multiple relation semantics. In detail, $\mathbf{\left(t-h\right)}$ is almost around only one translation vector $\mathbf{u_{r,m_{(h,r,t)}^*}}$ in TransG. Under the condition $m \neq m_{(h,r,t)}^*$, $\left( \mathbf{\frac{||\mathbf{u_h+u_{r,m}-u_t}||_2^2}{\sigma_h^2 + \sigma_t^2}} \right)$ is very large so that the exponential function value is very small. This is why the primary component could represent the corresponding semantics.

\begin{table}
	\centering
	\caption{Statistics of datasets}
	\label{tab1}
	\small
	\renewcommand\arraystretch{1.2}
	\begin{tabular}{|m{0.15\linewidth}<{\centering}|m{0.14\linewidth}<{\centering}|m{0.14\linewidth}<{\centering}|m{0.14\linewidth}<{\centering}|m{0.14\linewidth}<{\centering}|}
		\hline \textbf{Data} & \textbf{WN18} & \textbf{FB15K} & \textbf{WN11} & \textbf{FB13} \\
		\hline
		\hline \#Rel & 18 & 1,345 & 11 & 13 \\
		\hline \#Ent & 40,943 & 14,951 & 38,696 & 75,043 \\
		\hline \#Train & 141,442 & 483,142 & 112,581 & 316,232\\
		\hline \#Valid & 5,000 & 50,000 & 2,609 & 5,908\\
		\hline \#Test & 5,000 & 59,071 & 10,544 & 23,733\\
		\hline
	\end{tabular}
\end{table}

\begin{table*}
	\centering
	\caption{Evaluation results on link prediction}
	\label{tab2}
	\begin{tabular}{|*{9}{c|}}
		\hline \textbf{Datasets} & \multicolumn{4}{c|}{\textbf{WN18}}  &  \multicolumn{4}{c|}{\textbf{FB15K}}  \\
		\hline
		\hline
		\multirow{2}*{Metric} & \multicolumn{2}{c|}{Mean Rank} & \multicolumn{2}{c|}{HITS@10(\%)} & \multicolumn{2}{c|}{Mean Rank} & \multicolumn{2}{c|}{HITS@10(\%)} \\
		\cline{2-9} & Raw & Filter & Raw & Filter & Raw & Filter & Raw & Filter \\
		\hline
		Unstructured \cite{bordes2011learning} & 315 & 304 & 35.3 & 38.2 & 1,074 & 979 & 4.5 & 6.3 \\
		RESCAL \cite{nickel2012factorizing} & 1,180 & 1,163 & 37.2 & 52.8 & 828 & 683 & 28.4 & 44.1 \\
		SE　\cite{bordes2011learning} & 1,011 & 985 & 68.5 & 80.5 & 273 & 162 & 28.8 & 39.8  \\
		SME(bilinear) \cite{bordes2012joint} & 526 & 509 & 54.7 & 61.3 & 284 & 158 & 31.3 & 41.3  \\
		LFM \cite{jenatton2012latent} & 469 & 456 & 71.4 & 81.6 & 283 & 164 & 26.0 & 33.1  \\
		TransE \cite{bordes2013translating} & 263 & 251 & 75.4 & 89.2 & 243 & 125 & 34.9 & 47.1 \\
		TransH \cite{wang2014knowledge} & 401 & 388 & 73.0 & 82.3 & 212 & 87 & 45.7 & 64.4 \\
		TransR \cite{lin2015learning} & 238 & 225 & 79.8 & 92.0 & 198 & 77 & 48.2 & 68.7 \\
		CTransR \cite{lin2015learning} & \textbf{231} & \textbf{218} & 79.4 & 92.3 & 199 & 75 & 48.4 & 70.2 \\
		KG2E \cite{he2015learning} & 362 & 348 & 80.5 & 93.2 & \textbf{183} & \textbf{69} & 47.5 & 71.5 \\
		\hline \hline
		TransG (this paper) & 483 & 470 & \textbf{81.4} & \textbf{93.3} & 203 & 98 & \textbf{52.8} & \textbf{79.8} \\
		\hline
	\end{tabular}
\end{table*}

To summarize, previous studies make translation identically for all the triples of the same relation, but TransG automatically selects the best translation vector according to the specific semantics of a triple. Therefore, TransG could focus on the specific semantic embedding to avoid much noise from the other unrelated semantic components and result in promising improvements than existing methods. Note that, all the components in TransG have their own contributions, but the primary one makes the most.



\subsection{Training Algorithm}
The maximum data likelihood principle is applied for training. As to the non-parametric part, $\pi_{r,m}$ is generated from the CRP with Gibbs Sampling, similar to \cite{he2015learning} and \cite{griffiths2011indian}. A new component is sampled for a triple (h,r,t) with the below probability:
\begin{eqnarray}
\mathbb{P}(m_{r, new}) = \frac{
	\beta e^
	{
		- \frac
		{||\mathbf{h-t}||_2^2}
		{\sigma_h^2 + \sigma_t^2 + 2}
	}
}
{
	\beta
	e^{
		- \frac
		{||\mathbf{h-t}||_2^2}
		{\sigma_h^2 + \sigma_t^2 + 2}
	} + \mathbb{P}\{(h, r, t)\}
}
\end{eqnarray}

\begin{table*}
	\centering
	\caption{Evaluation results on FB15K by mapping properties of relations(\%)}
	\label{tab3}
	\begin{tabular}{|*{9}{c|}}
		\hline \textbf{Tasks} & \multicolumn{4}{c|}{\textbf{Predicting Head(HITS@10)}}  &  \multicolumn{4}{c|}{\textbf{Predicting Tail(HITS@10)}}  \\
		\hline
		\hline
		Relation Category & 1-1 & 1-N & N-1 & N-N & 1-1 & 1-N & N-1 & N-N \\
		\hline
		Unstructured \cite{bordes2011learning} & 34.5 & 2.5 & 6.1 & 6.6 & 34.3 & 4.2 & 1.9 & 6.6 \\
		SE　\cite{bordes2011learning} & 35.6 & 62.6 &17.2 & 37.5& 34.9 & 14.6 & 68.3 & 41.3  \\
		SME(bilinear) \cite{bordes2012joint} & 30.9 & 69.6 & 19.9 & 38.6 & 28.2 & 13.1 & 76.0 & 41.8 \\
		TransE \cite{bordes2013translating} & 43.7 &65.7 & 18.2 & 47.2 & 43.7 & 19.7 & 66.7 & 50.0 \\
		TransH \cite{wang2014knowledge} & 66.8 & 87.6 & 28.7 & 64.5 & 65.5 & 39.8 & 83.3 & 67.2 \\
		TransR \cite{lin2015learning} & 78.8 & 89.2 & 34.1 & 69.2 & 79.2 & 37.4 & 90.4 & 72.1 \\
		CTransR \cite{lin2015learning} & 81.5 & 89.0 & 34.7 & 71.2 & 80.8 & 38.6 & 90.1 & 73.8 \\
		\hline \hline
		TransG (this paper) & \textbf{85.4} & \textbf{95.7} & \textbf{44.7} & \textbf{80.8} & \textbf{84.0} &
		\textbf{56.5} & \textbf{95.0} & \textbf{83.3} \\
		\hline
	\end{tabular}
\end{table*}

where $\mathbb{P}\{(h, r, t)\}$ is the current posterior probability. As to other parts, in order to better distinguish the true triples from the false ones, we maximize the ratio of likelihood of the true triples to that of the false ones. Notably, the embedding vectors are initialized by \cite{glorot2010understanding}. Putting all the other constraints together, the final objective function is obtained, as follows:
\begin{eqnarray}
& & \min_{\mathbf{u_h, u_{r,m}, u_t}} \mathcal{L} \nonumber \\
& & \mathcal{L} = -\sum_{(h,r,t)\in \Delta} ln \left( \sum_{m=1}^{M_r} \pi_{r,m} e^{-\frac{||\mathbf{u_h+u_{r,m}-u_t}||_2^2}{\sigma_h^2 + \sigma_t^2}} \right) \nonumber \\
& & + \sum_{(h',r',t')\in \Delta'} ln \left( \sum_{m=1}^{M_r} \pi_{r',m} e^{-\frac{||\mathbf{u_{h'} + u_{r',m}-u_{t'}}||_2^2}{\sigma_{h'}^2 + \sigma_{t'}^2}} \right) \nonumber \\
& & + C\left( \sum_{r\in R} \sum_{m=1}^{M_r} ||\mathbf{u_{r,m}}||_2^2 + \sum_{e \in E}||\mathbf{u_e}||^2_2  \right) 
\end{eqnarray}
where $\Delta$ is the set of golden triples and $\Delta'$ is the set of false triples. $C$ controls the scaling degree. $E$ is the set of entities and $R$ is the set of relations. Noted that the mixing factors $\pi$ and the variances $\sigma$ are also learned jointly in the optimization. 

SGD is applied to solve this optimization problem. In addition, we apply a trick to control the parameter updating process during training. For those very impossible triples, the update process is skipped. Hence, we introduce a similar condition as TransE \cite{bordes2013translating} adopts: the training algorithm will update the embedding vectors only if the below condition is satisfied:
\begin{eqnarray}
\frac{\mathbb{P}\{(h,r,t)\}}{\mathbb{P}\{(h',r',t')\}} & = & \frac {\sum_{m=1}^{M_r} \pi_{r,m} e^{-\frac{||\mathbf{u_h+u_{r,m}-u_t}||_2^2}{\sigma_h^2 + \sigma_t^2}}} {\sum_{m=1}^{M_{r'}} \pi_{r',m} e^{-\frac{||\mathbf{u_{h'} + u_{r',m}-u_{t'}||_2^2}}{\sigma_{h'}^2 + \sigma_{t'}^2}}} \nonumber \\
& \le & M_r e^\gamma
\end{eqnarray}
where $(h,r,t) \in \Delta$ and $(h',r',t') \in \Delta'$. $\gamma$ controls the updating condition.

As to the efficiency, in theory, the time complexity of TransG is bounded by a small constant M compared to TransE, that is $O(\mathrm{TransG}) = O(M \times O(\mathrm{TransE}))$ where M is the number of semantic components in the model. Note that TransE is the fastest method among translation-based methods. The experiment of Link Prediction shows that TransG and TransE would converge at around 500 epochs, meaning there is also no significant difference in convergence speed. In experiment, TransG takes 4.8s for one iteration on FB15K while TransR costs 136.8s and PTransE costs 1200.0s on the same computer for the same dataset.

\section{Experiments}
Our experiments are conducted on four public benchmark datasets that are the subsets of Wordnet and Freebase, respectively. The statistics of these datasets are listed in Tab.\ref{tab1}. Experiments are conducted on two tasks : Link Prediction and Triple Classification. To further demonstrate how the proposed model approaches multiple relation semantics, we present semantic component analysis at the end of this section.

\subsection{Link Prediction}
Link prediction concerns knowledge graph completion: when given an entity and a relation, the embedding models predict the other missing entity. More specifically, in this task, we predict $t$ given $(h, r, *)$, or predict $h$ given $(*,  r, t)$. The WN18 and FB15K are two benchmark datasets for this task. Note that many AI tasks could be enhanced by Link Prediction such as relation extraction \cite{hoffmann2011knowledge}.

\begin{figure*}
	\centering
	\includegraphics[width=0.75\linewidth]{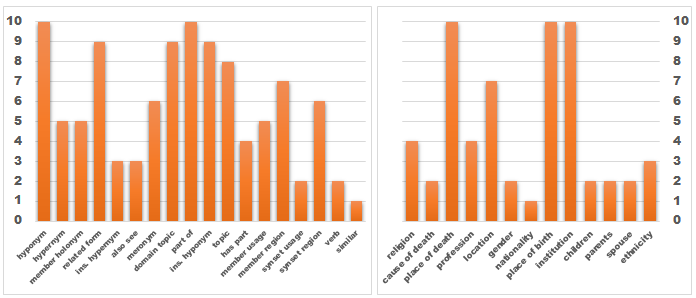}
	\caption{Semantic component number on WN18 (left) and FB13 (right).}
	\label{fig:fig_5}
\end{figure*}

\begin{table*}
	\centering
	\caption{Different clusters in WN11 and FB13 relations.}
	\label{tab5}
	\begin{tabular}{c|c|c}
		\hline  \textbf{Relation} & \textbf{Cluster} & \textbf{Triples (Head, Tail)} \\
		\hline \multirow{2}{*}{$\mathrm{PartOf}$} & Location &  (Capital of Utah, Beehive State), (Hindustan, Bharat) ...\\
		\cline{2-3} & Composition &  (Monitor, Television), (Bush, Adult Body), (Cell Organ, Cell)... \\
		\hline \multirow{2}{*}{$\mathrm{Religion}$} & Catholicism & (Cimabue, Catholicism), (St.Catald, Catholicism) ... \\
		\cline{2-3} & Others & (Michal Czajkowsk, Islam), (Honinbo Sansa, Buddhism) ... \\
		\hline \multirow{2}{*}{$\mathrm{DomainRegion}$} & Abstract & (Computer Science, Security System), (Computer Science, PL).. \\
		\cline{2-3} & Specific & (Computer Science, Router), (Computer Science, Disk File) ... \\
		\hline  \multirow{3}{*}{$\mathrm{Profession}$} & Scientist & (Michael Woodruf, Surgeon), (El Lissitzky, Architect)... \\
		\cline{2-3} & Businessman& (Enoch Pratt, Entrepreneur), (Charles Tennant, Magnate)... \\
		\cline{2-3} & Writer & (Vlad. Gardin, Screen Writer), (John Huston, Screen Writer) ...\\
		\hline
	\end{tabular}
\end{table*}

\textbf{Evaluation Protocol.} We adopt the same protocol used in previous studies. For each testing triple $(h,r,t)$, we corrupt it by replacing the tail $t$ (or the head $h$) with every entity $e$ in the knowledge graph and calculate a probabilistic score of this corrupted triple $(h,r,e)$ (or $(e,r,t)$) with the score function $f_r(h,e)$. After ranking these scores in descending order, we obtain the rank of the original triple. There are two metrics for evaluation: the averaged rank (Mean Rank) and the proportion of testing triple whose rank is not larger than 10 (HITS@10). This is called ``Raw'' setting. When we filter out the corrupted triples that exist in the training, validation, or test datasets, this is the``Filter'' setting. If a corrupted triple exists in the knowledge graph, ranking it ahead the original triple is also acceptable. To eliminate this case, the ``Filter'' setting is preferred. In both settings, a lower Mean Rank and a higher HITS@10 mean better performance.

\textbf{Implementation.} As the datasets are the same, we directly report the experimental results of several baselines from the literature, as in \cite{bordes2013translating}, \cite{wang2014knowledge} and \cite{lin2015learning}. We have attempted several settings on the validation dataset to get the best configuration. For example, we have tried the dimensions of {100, 200, 300, 400}. Under the ``bern.'' sampling strategy, the optimal configurations are: learning rate $\alpha = 0.001$, the embedding dimension $k=100$, $\gamma=2.5$, $\beta=0.05$ on WN18; $\alpha=0.0015$, $k=400$, $\gamma=3.0$, $\beta=0.1$ on FB15K. Note that all the symbols are introduced in ``Methods''. \textit{We train the model until it converges in previous version (about 10,000 rounds), but we provide the results of 2,000 rounds for comparison in current version.}

\textbf{Results.} Evaluation results on WN18 and FB15K are reported in Tab.\ref{tab2} and Tab.\ref{tab3}. We observe that:
\begin{enumerate}
	\item TransG outperforms all the baselines obviously. Compared to TransR, TransG makes improvements by 2.9\% on WN18 and 26.0\% on FB15K, and the averaged semantic component number on WN18 is 5.67 and that on FB15K is 8.77. This result demonstrates capturing multiple relation semantics would benefit embedding.
	\item The model has a bad Mean Rank score on the WN18 and FB15K dataset. Further analysis shows that there are 24 testing triples (0.5\% of the testing set) whose ranks are more than 30,000, and these few cases would lead to about 150 mean rank loss. Among these triples, there are 23 triples whose tail or head entities have never been co-occurring with the corresponding relations in the training set. In one word, there is no sufficient training data for those relations and entities. 
	\item Compared to CTransR, TransG solves the multiple relation semantics problem much better for two reasons. Firstly, CTransR clusters the entity pairs for a specific relation and then performs embedding for each cluster, but TransG deals with embedding and multiple relation semantics simultaneously, where the two processes can be enhanced by each other. Secondly, CTransR models a triple by only one cluster, but TransG applies a mixture to refine the embedding. 
\end{enumerate}
Our model is almost insensitive to the dimension if that is sufficient. For the dimensions of ${100, 200, 300, 400}$, the HITS@10 of TransG on FB15 are ${81.8\%, 84.0\%, 85.8\%, 88.2\%}$, while those of TransE are ${47.1\%, 48.5\%, 51.3\%, 49.2\%}$.

\subsection{Triple Classification}
In order to testify the discriminative capability between true and false facts, triple classification is conducted. This is a classical task in knowledge base embedding, which aims at predicting whether a given triple $(h,r, t)$ is correct or not. WN11 and FB13 are the benchmark datasets for this task. Note that evaluation of classification needs negative samples, and the datasets have already provided negative triples.

\textbf{Evaluation Protocol.} The decision process is very simple as follows: for a triple $(h,r,t)$, if $f_r(h,t)$ is below a threshold $\sigma_r$, then positive; otherwise negative. The thresholds $\{\sigma_r\}$ are determined on the validation dataset.

\begin{figure}
	\centering
	\includegraphics[width=1.0\linewidth]{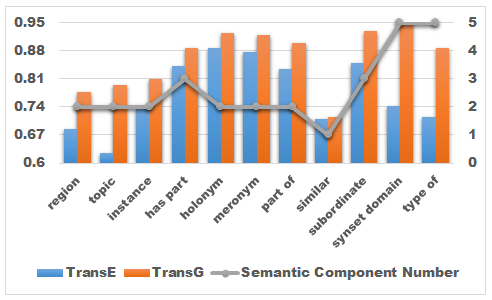}
	\caption{Accuracies of each relations in WN11 for triple classification. The right y-axis is the number of semantic components, corresponding to the lines.}
	\label{fig:fig_4_1}
\end{figure}

\begin{table}[H]
	\centering
	\caption{Triple classification: accuracy(\%) for different embedding methods.}
	\label{tab4}
	\begin{tabular}{|c|c|c|c|}
		\hline \textbf{Methods} & \textbf{WN11} & \textbf{FB13} & \textbf{AVG.} \\
		\hline
		LFM  & 73.8 & 84.3 & 79.0\\
		NTN  & 70.4 & 87.1 & 78.8 \\
		TransE & 75.9 & 81.5 & 78.7 \\
		TransH & 78.8 & 83.3 & 81.1 \\
		TransR & 85.9 & 82.5 & 84.2\\
		CTransR & 85.7 & N/A & N/A \\
		KG2E & 85.4 & 85.3 & 85.4 \\
		\hline \hline
		TransG & \textbf{87.4} & \textbf{87.3} & \textbf{87.4} \\
		\hline
	\end{tabular}
\end{table}

\textbf{Implementation.} As all methods use the same datasets,  we directly re-use the results of different methods from the literature. We have attempted several settings on the validation dataset to find the best configuration. The optimal configurations of TransG are as follows: ``bern'' sampling, learning rate $\alpha= 0.001$, $k=50$, $\gamma=6.0$, $\beta=0.1$ on WN11, and ``bern'' sampling, $\alpha=0.002$, $k=400$,  $\gamma=3.0$, $\beta=0.1$ on FB13. 

\textbf{Results.} Accuracies are reported in Tab.\ref{tab4} and Fig.\ref{fig:fig_4_1}. The following are our observations:
\begin{enumerate}
	\item TransG outperforms all the baselines remarkably. Compared to TransR, TransG improves by 1.7\% on WN11 and 5.8\% on FB13, and the averaged semantic component number on WN11 is 2.63 and that on FB13 is 4.53. This result shows the benefit of capturing multiple relation semantics for a relation.
	\item The relations, such as ``Synset Domain'' and ``Type Of'', which hold more semantic components, are improved much more. In comparison, the relation ``Similar'' holds only one semantic component and is almost not promoted. This further demonstrates that capturing multiple relation semantics can benefit embedding.
\end{enumerate}

\subsection{Semantic Component Analysis}
In this subsection, we analyse the number of semantic components for different relations and list the component number on the dataset WN18 and FB13 in Fig.\ref{fig:fig_5}.

\textbf{Results.} As Fig.~\ref{fig:fig_5} and Tab.~\ref{tab5} show, we have the following observations:
\begin{enumerate}
	\item Multiple semantic components are indeed necessary for most relations. Except for relations such as ``Also See'', ``Synset Usage'' and ``Gender'', all other relations have more than one semantic component.
	\item Different components indeed correspond to different semantics, justifying the theoretical analysis and effectiveness of TransG. For example, ``Profession'' has at least three semantics: scientist-related as $\mathrm{(El Lissitzky, Architect)}$, businessman-related as $\mathrm{(Enoch Pratt, Entrepreneur)}$ and writer-related as $\mathrm{(Vlad. Gardin, Screen Writer)}$.
	\item WN11 and WN18 are different subsets of Wordnet. As we know, the semantic component number is decided on the triples in the dataset. Therefore, It's reasonable that similar relations, such as ``Synset  Domain'' and ``Synset Usage'' may hold different semantic numbers for WN11 and WN18.
\end{enumerate}

\section{Conclusion}
In this paper, we propose a generative Bayesian non-parametric infinite mixture embedding model, TransG, to address a new issue, multiple relation semantics, which can be commonly seen in knowledge graph. TransG can discover the latent semantics of a relation automatically and leverage a mixture of relation components for embedding. Extensive experiments show our method achieves substantial improvements against the state-of-the-art baselines.

\textbf{Support Materials.}
All of the related poster, slides, datasets and codes \textbf{\em HAVE BEEN ALREADY} published in \url{http://www.ibookman.net/conference.html}.

\section{Code Tricks}
	\begin{enumerate}
	\item The class ``TransG'' is the experimental version, rather than ``TransG\_Hierarchical''. Note that, for numerical stabilization, we fix the variance $\sigma$ as a constant.
	\item When generating the new cluster, we assign the CRP factor as the mixture factor, which is required by CRP theoretically. But as to the center of new cluster, we assign a random vector rather than $\mathbf{t-h}$. Because in big data scenario, theoretical center is far away from the ground-truth.
	\item Regarding the learning methodology of parameter $\pi$ and $\sigma$, we applied the stochastic gradient ascent (SGD) for efficiency, rather than likelihood counting, which is naturally suitable for CRP but inefficient. We suggest the reader to implement both for experiments.
	\item Due to the slow convergence rate, in previous version, we train the model until convergence, almost around 10,000 epos. But in current version, we conduct the same rounds with other baselines （2,000 rounds).
\end{enumerate}


\bibliography{acl}
\bibliographystyle{acl}

\end{document}